\documentclass[10pt,twocolumn,letterpaper]{article}

\usepackage{cvpr}              
\definecolor{cvprblue}{rgb}{0.21,0.49,0.74}
\usepackage[pagebackref,breaklinks,colorlinks,allcolors=cvprblue,pdfborder={0 0 0}]{hyperref}
\definecolor{c2}{HTML}{FBD9BD}
\definecolor{c3}{HTML}{fe793d}
\definecolor{c4}{HTML}{eedeb0}
\definecolor{pp}{HTML}{BC7FCD}
\definecolor{bb}{HTML}{CDE8E5}
\definecolor{rouse}{rgb}{0.981,0.961,0.941}
\usepackage{makecell}
\usepackage{xcolor}
\usepackage{multirow}
\usepackage{booktabs}
\usepackage{colortbl} 
\usepackage{wrapfig}
\usepackage[table]{xcolor}
\usepackage{ulem}
\usepackage{url}
\def\paperID{4820} 
\def\confName{CVPR}
\def\confYear{2026}

\title{Beyond Ground-Truth: Leveraging Image Quality Priors for \\ Real-World Image Restoration}

\author{Fengyang Xiao$^{1,*}$\,,
Peng Hu$^{2,*}$\,,
Lei Xu$^{3}$\,,        
XingE Guo$^{1}$\,,
Guanyi Qin$^{1}$\,,
        Yuqi Shen$^{2}$\,,
          \\
          Chengyu Fang$^{2}$\,,
          Rihan Zhang$^{1}$\,,
            Chunming He$^{1,\dagger}$\,, 
        and {Sina Farsiu}$^{1}$\\
        $^1$Duke University,
	$^2$Tsinghua University,  $^3$EPFL 
\\
$*$ Equal Contribution, $\dagger$ Corresponding Author, \\
Contact: fengyang.xiao@duke.edu / tommypinkman47@gmail.com / chunming.he@duke.edu
}

\begin{document}

\twocolumn[{
\maketitle
\vspace{-12mm}
\begin{center}
\includegraphics[width=\textwidth]{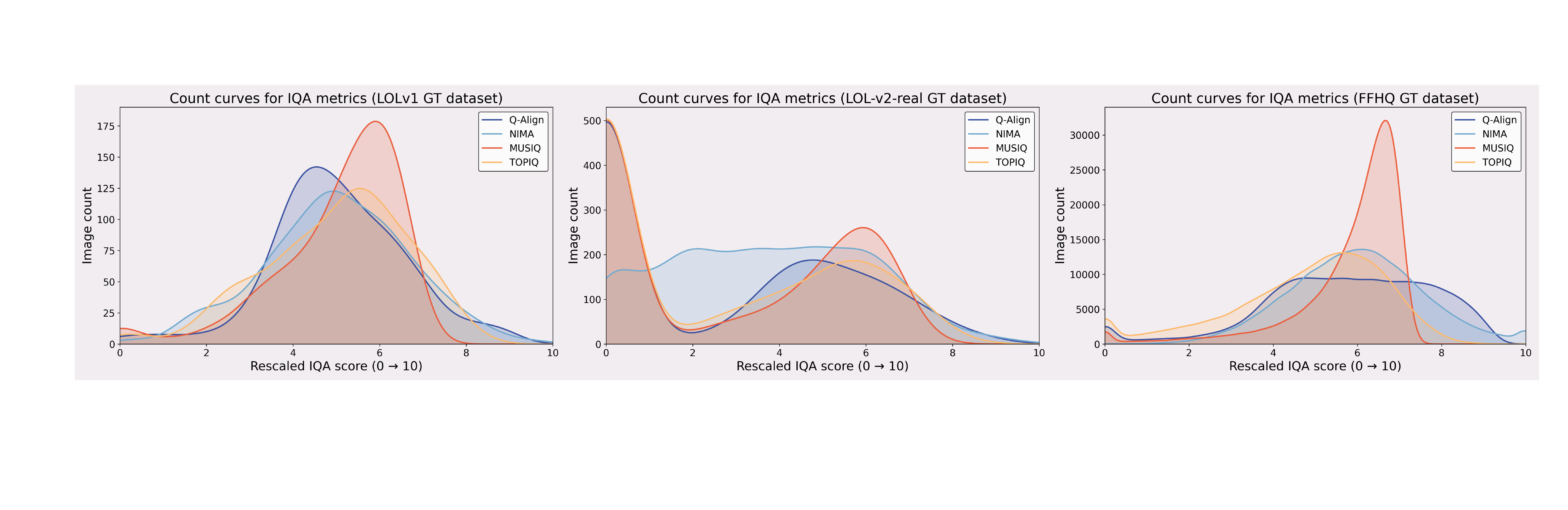}\vspace{-3mm} 
\captionof{figure}{{Score distributions given by different Image Quality Assessment (IQA) models across various ground truth datasets, where higher scores indicate better image quality. Notably, most ground-truth images exhibit an average quality score between 5 and 8, with some residual degradation patterns. This limits the network’s ability to output truly high-quality images, achieving scores closer to 9.}} \label{Fig:score_distribution}
\end{center}
}]

\begin{abstract}
Real-world image restoration aims to restore high-quality (HQ) images from degraded low-quality (LQ) inputs captured under uncontrolled conditions. Existing methods typically depend on ground-truth (GT) supervision, assuming that GT provides perfect reference quality. However, GT can still contain images with inconsistent perceptual fidelity, causing models to converge to the average quality level of the training data rather than achieving the highest perceptual quality attainable.
To address these problems, we propose a novel framework, termed \textbf{IQPIR}, that introduces an Image Quality Prior (IQP)—extracted from pre-trained No-Reference Image Quality Assessment (NR-IQA) models—to guide the restoration process toward perceptually optimal outputs explicitly. Our approach synergistically integrates IQP with a learned codebook prior through three key mechanisms:
(1) a \textbf{quality-conditioned Transformer}, where NR-IQA-derived scores serve as conditioning signals to steer the predicted representation toward maximal perceptual quality. This design provides a plug-and-play enhancement compatible with existing restoration architectures without structural modification; and
(2) a \textbf{dual-branch codebook structure}, which disentangles common and HQ-specific features, ensuring a comprehensive representation of both generic structural information and quality-sensitive attributes; and (3) {a \textbf{discrete representation-based quality optimization strategy}, which mitigates over-optimization effects commonly observed in continuous latent spaces.}
Extensive experiments on real-world image restoration demonstrate that our method not only surpasses cutting-edge methods but also serves as a generalizable quality-guided enhancement strategy for existing methods. 
The code is available at \url{https://github.com/fengyang1399-pixel/IQPIR}.

\end{abstract}

\section{Introduction}\label{sec:intro}
Image restoration aims to reconstruct degraded low-quality (LQ) images into high-quality (HQ) outputs, yet real-world degradations are complex and unknown, making the task highly ill-posed.
Recent works leverage diverse priors to enrich details and enhance robustness under various degradations\cite{ma2020deep,li2020blind,li2020enhanced,wang2021gfpgan,wang2023dr2,IDM}.
Among them, codebook-based methods\cite{codeformer,gu2022vqfr,wang2022restoreformer,tsai2023dual} formulate restoration as a code prediction problem in a discrete representation space, effectively reducing reconstruction ambiguity and improving resilience to degradation diversity.

\begin{figure}[htbp]   \includegraphics[scale=0.3]{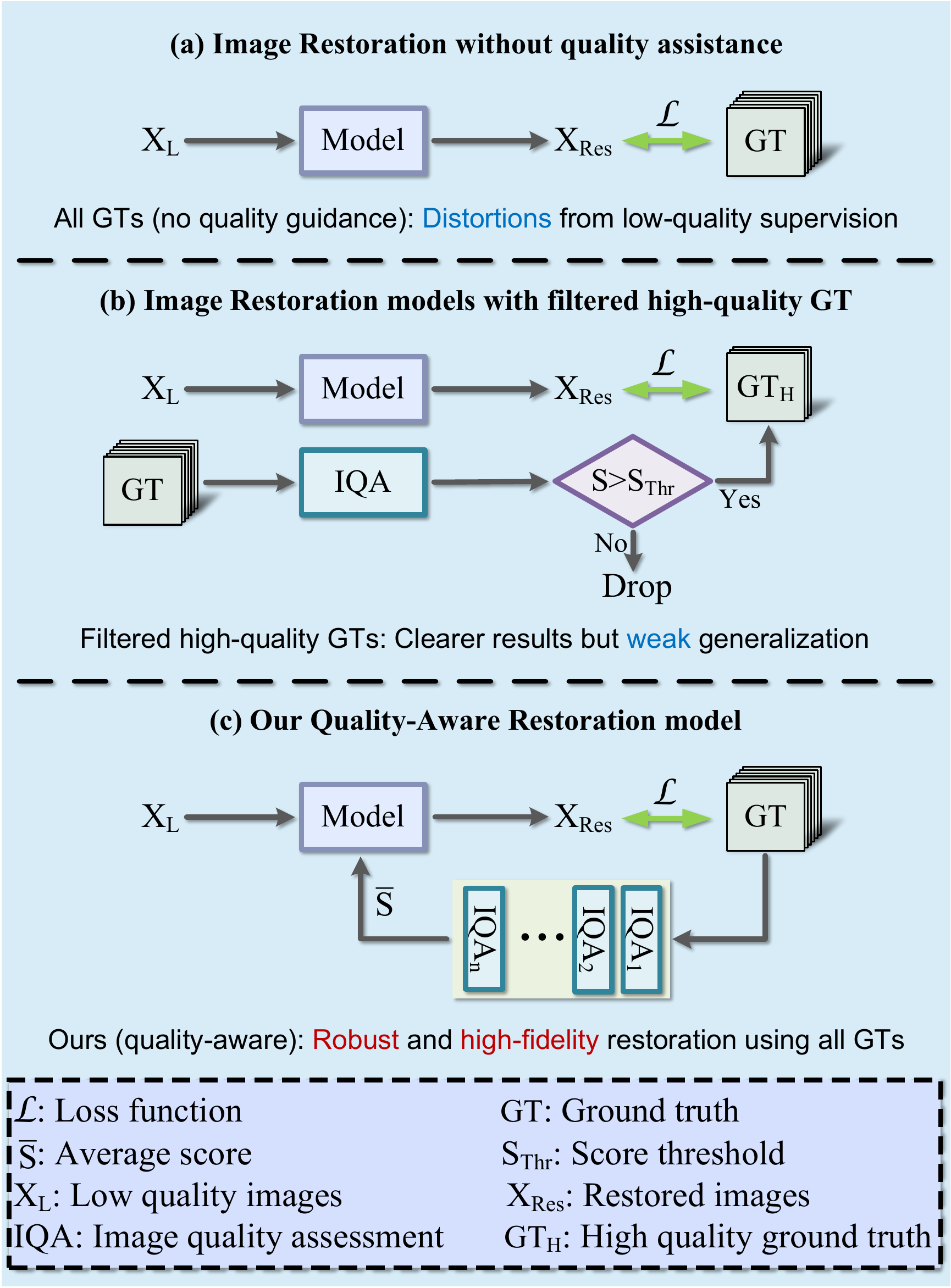}\setlength{\abovecaptionskip}{-0cm}
 \vspace{-0.3cm}
  \caption{Structural comparison of different training paradigms, where both (b) and our (c) utilize IQA to guide network training.}
    \label{Fig:three_models}
\vspace{-0.6cm}
\end{figure}

However, a basic limitation of existing methods lies in the implicit assumption that high-quality ground truth (GT) data are flawless and should serve as the sole supervision source. As shown in \cref{Fig:score_distribution}, the perceptual quality within these GT datasets is often inconsistent and suboptimal. Consequently, most models restore LQ inputs only to match the average GT quality, rather than pursuing the highest achievable perceptual fidelity. Although employing NR-IQA models to filter out and retain only the highest-quality (HQ+) images can improve overall output quality, this strategy inevitably reduces data diversity, as top-tier GT samples constitute only a small subset, as shown in \cref{Fig:three_models}. Over-reliance on such curated data may lead to artifacts and degraded feature representations in the restored outputs.

\begin{figure}[tbp]
  \setlength{\abovecaptionskip}{-0.25cm}
    \includegraphics[width=\linewidth,scale=0.1]{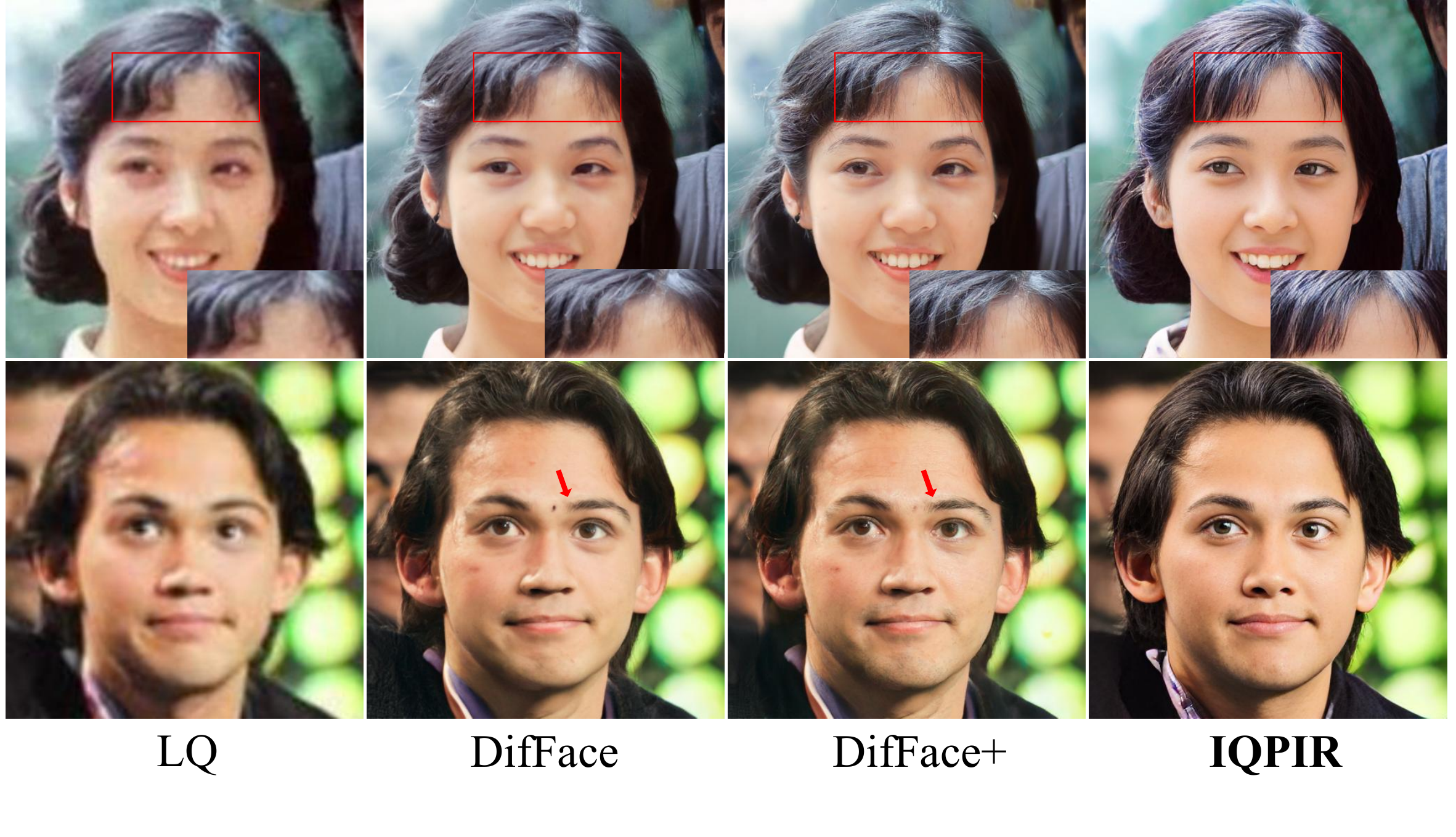}
    \caption{Results on blind face restoration. DifFace+ is the DifFace~\citep{yue2022difface} with our quality prior conditioned approach, having more details. Our IQPIR has the highest perceptual quality.}
    \label{Fig:first_show}
\vspace{-0.6cm}
\end{figure}

To overcome this, we propose a novel framework that uses an Image Quality Prior to address the suboptimal restoration quality caused by imperfect GT. Our IQP is derived from pre-trained Non-Reference Image Quality Assessment (NR-IQA) models, which enables our restoration model to distinguish between different quality levels and associate images with their corresponding quality scores. Instead of blindly pursuing a match to imperfect GT, our core idea is to explicitly guide the restoration process toward the highest possible perceptual quality. As shown in~\cref{Fig:first_show}, our strategy can also facilitate cutting-edge methods.

We introduce a novel framework, IQPIR, that seamlessly integrates image quality and codebook priors to enhance restoration quality. We first obtain an image quality prior from NR-IQA models and incorporate it at different training stages. In the code prediction stage, we use a \textbf{quality-conditioned Transformer} that takes the quality score as a condition during code prediction. This allows the model to generate images with the highest quality when provided with the highest score as a condition. The feasibility of this score-conditioned manner stems from the inconsistent quality of training data, making it applicable to other restoration models.
We also leverage the discrete representation space of the codebook for quality optimization, which mitigates the over-optimization issue common in continuous representations. 
{Furthermore, different GT quality tiers offer distinct functional advantages: High-Quality (HQ) GT excels in controlling fine-grained structures (e.g.,  hair-end details), while average GT is more effective for generalized recovery from casual, large-area distortions (e.g., blurs). Based on this principle,} in the codebook prior learning stage, we introduce a \textbf{dual-codebook architecture}, comprising a common codebook and an HQ+ codebook, to learn both general and high-quality-specific features.
Together, these innovations enable robust and perceptually consistent restoration across diverse real-world conditions.

In summary, our main contributions are as follows:

\noindent(1) We propose a novel prior-based method, IQPIR, which uses an image quality prior to help restoration models understand HQ features and improves the ability to generate HQ results. 
This bridges the gap between an image quality prior and its application in enhancing restoration quality.

\noindent(2) We introduce a series of strategies for integrating image quality prior with codebook prior, including a plug-and-play score-conditioned approach, a quality optimization technique based on discrete representation, and a dual-codebook reconstruction model. These strategies leverage the advantages of the image quality prior across multiple stages to enhance restoration quality.

\noindent(3) Experiments demonstrate that IQPIR effectively generalizes across various real-world restoration tasks, consistently delivering superior perceptual quality and robustness compared to existing state-of-the-art methods.

\begin{figure*}[!htbp]
  \centering
    \setlength{\abovecaptionskip}{-0.0cm}
      \vspace{-0.4cm}
  \includegraphics[width=\linewidth]{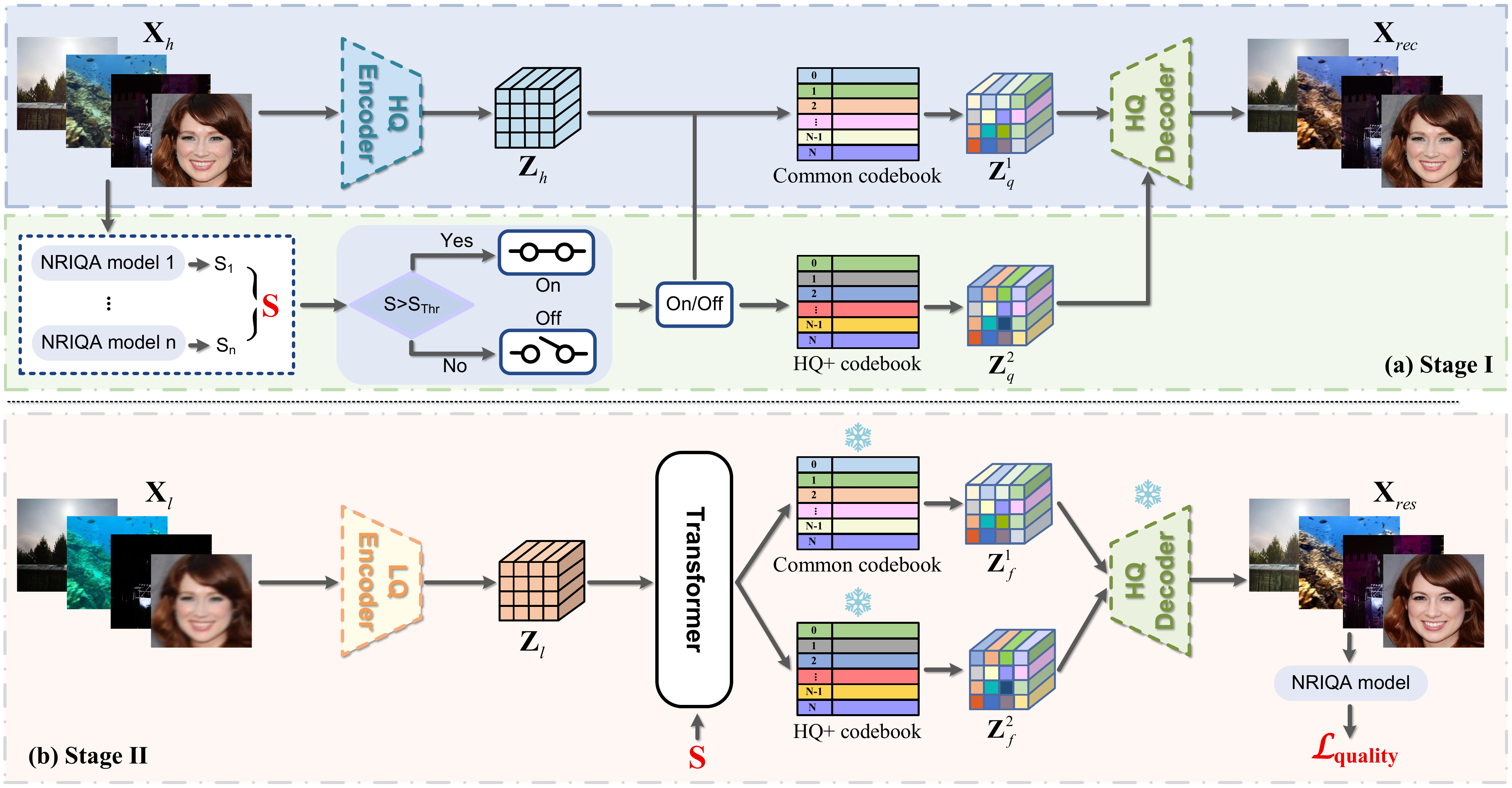}
  \caption{\textbf{Overall framework of IQPIR.} (a) In the codebook learning stage, a dual-codebook architecture is proposed. The HQ+ codebook is learned to quantize $Z_h$ only when the quality score of $x_h$ is higher than the threshold $S_{thr}$.
  (b) In the codebook lookup stage, we input the quality score $S$ as a condition into Transformer \textit{T}, which predicts two code sequences at the same time. 
  The two codebooks are leveraged to look up the corresponding code entries. Finally, the NR-IQA model is utilized to calculate the quality loss $\mathcal{L}_{quality}$.
  }
  \vspace{-0.4cm}
  \label{fig:architecture}
\end{figure*}
\section{Related Work}
\label{sec: relatedwork}
\noindent \textbf{Real-world image restoration}. RWIR aims to recover HQ images from degraded LQ inputs affected by authentic noise, blur, or illumination variations~\cite{he2025reti,qin2023data,deng2022pcgan,he2025segment,he2025run}. 
Early works relied on explicit degradation modeling~\cite{he2025reti} or paired data synthesis~\cite{yue2020dual}, but limited by inaccurate assumptions. 
Later approaches leveraged large-scale datasets~\cite{yu2024scaling} or degradation-aware architectures~\cite{he2023degradation} to better approximate real-world conditions. 
Recent methods, such as domain adaptation~\cite{fang2024real}, unpaired learning~\cite{he2023hqg}, and self-supervised objectives~\cite{he2025unfoldir}, further reduce the gap between synthetic and real data.
However, most methods still assume that GT images are perfect, even though datasets like FFHQ~\cite{karras2019style} exhibit uneven perceptual quality, leading models to converge toward average GT fidelity rather than optimal perceptual quality. Our IQPIR introduces an image quality prior derived from NR-IQA models to guide training toward higher-quality reconstructions under diverse degradations.

\noindent \textbf{Prior-based image restoration}.
Priors play a crucial role in restoration problems. Common priors include geometric~\cite{ma2020deep,hu2021face}, reference~\cite{gfrnet,li2020blind,li2020enhanced}, generative~\cite{wang2021gfpgan,he2022gcfsr,wang2023dr2,IDM,he2025reversible}, and codebook priors~\cite{gu2022vqfr,codeformer,tsai2023dual}. 
Geometric priors provide structural cues~\cite{hu2021face,chen2021progressive} but degrade under heavy distortions. Reference priors \cite{li2020blind,lin2023diffbir} rely on extra HQ data, limiting applicability. Generative priors \cite{wang2023dr2,IDM,li2025interlcm} exploit pretrained models for semantic completion but may hallucinate. Codebook priors \cite{gu2022vqfr,codeformer,tsai2023dual}, based on vector quantization \cite{vqgan}, stabilize restoration by learning discrete representations, achieving perceptual gains.
Yet, they assume flawless GT supervision, overlooking quality inconsistency within training data.
Our IQPIR extends codebook-based methods with an image quality prior as a perception-aware constraint, enabling the network to adaptively emphasize high-quality cues and achieve high-fidelity restoration.

A preliminary version of our core idea, which uses NR-IQA scores as quality priors to guide restoration, appeared in an earlier preprint~\cite{hu2025iqpfr}. The present work substantially extends that effort by introducing a unified dual-codebook framework, discrete-space quality optimization, and a comprehensive evaluation across diverse restoration tasks.

\setlength{\abovedisplayskip}{2pt}
\setlength{\belowdisplayskip}{2pt}

\section{Methodology}
To address the challenge of real-world image restoration (RWIR), where degradations are complex and annotations imperfect, we introduce IQPIR, a unified framework that incorporates Image Quality Priors (IQPs) into both the codebook learning and code prediction stages. The pipeline is shown in~\cref{fig:architecture}. IQPIR integrates perceptual quality awareness into the discrete representation learning process to produce outputs of the highest attainable fidelity and realism.

In the first stage, a dual-codebook architecture is trained to disentangle universal structural features from high-quality (HQ+) attributes (\cref{subsec:3.1}). In the second stage, a quality-prior-conditioned transformer exploits the learned dual codebook for restoration, with image quality priors integrated both as input conditions and as optimization objectives (\cref{3.2,3.3}).

\subsection{Dual Codebook Learning}
\label{subsec:3.1}
Real-world images exhibit highly variable perceptual quality. To robustly represent both common structures and high-quality details, we employ a dual codebook design with a common codebook and a high-quality (HQ+) codebook.

During training, HQ images $x_h$ are first encoded by the feature extractor $E$ and quantized via the common codebook to obtain $Z_q^1$. If a given image’s quality score $S$, estimated using a No-Reference Image Quality Assessment (NR-IQA) model, exceeds a threshold $S_{thr}$, its features are further quantized using the HQ+ codebook to generate $Z_q^2$. The final quantized feature is defined as
\begin{equation}
  Z_q=
  \begin{cases}
    Z_q^1+\alpha Z_q^2  & if \text{ } S>S_{thr} ,  \\
    Z_q^1 & if \text{ } S <= S_{thr}.
  \end{cases}
  \label{eq:dual_codebook}
\end{equation}
where $\alpha$ balances the contributions of the two codebooks. The decoder $\textit{D}$ then reconstructs the image $x_{rec}$ from $Z_q$.

The common codebook learns generic structures across diverse training data, while the HQ+ codebook specializes in the fine-grained visual details of high-quality samples. This separation enables the model to preserve general fidelity while enhancing perceptual quality. At inference, both codebooks are jointly leveraged, allowing the model to synthesize HQ+ outputs even from degraded inputs.

\subsection{Quality Prior Condition}
\label{3.2}
Before training, we first employ NR-IQA models to assess the quality score of GT data, denoted as $S$. In the codebook lookup stage, we introduce the quality prior as an additional condition for code prediction using a conditional Transformer. This Transformer model predicts two code sequences conditioned on the image quality score $S$. The conditional Transformer is trained to capture the relationship between GT images and their associated quality scores. Specifically, during training, when the highest quality score is used as the condition, the ground truth is set to the highest quality image. This strategy resembles class-conditional generation, enabling the model to learn quality-specific generation capabilities across diverse levels.

First, we extract the LQ feature via $Z_l = \textit{E}(x_l)$. 
Then we embed the quality score $S$ into an embedding vector $\textbf{s} \in \mathbb{R}^{h*w*c} $ and reshape it to $\textbf{s} \in \mathbb{R}^{h \times w \times c} $, which matches the dimensionality of $Z_l$. 
This embedding vector $\textbf{s}$ is directly added to $Z_l$, formulated as:
\begin{equation}
  \centering
  \begin{aligned}
      \widehat{Z}_l=Z_l+\textbf{s}.
    \label{eq:z1_z2}
  \end{aligned}
\end{equation}
Upon receiving $\widehat{Z}_l$ as input, the Transformer $\textit{T}$ predicts two code sequences, $\textbf{c}_1$ and $\textbf{c}_2$.
Then $\textbf{c}_1$ retrieves code entries from the common codebook to construct the quantized feature $Z_f^1$, while $\textbf{c}_2$ retrieves code entries from the HQ+ codebook to form  $Z_f^2$.
We then fuse $Z_f^1$ and $Z_f^2$ to get $Z_f$ via: 
\begin{equation}
  \centering
  \begin{aligned}
      Z_f=Z_f^1+\alpha Z_f^2,
    \label{eq:zf}
  \end{aligned}
\end{equation}
where $\alpha$ is the balance weight. $Z_f$ is fed to the decoder to generate restoration images $x_{res}$. Since \textit{D} is trained to reconstruct high-quality (HQ+) images from the fused representations $Z_q^1$ and $Z_q^2$, it is capable of generating restorations with comparable perceptual fidelity when taking the fused feature $Z_f$ as input during the second stage.

By conditioning on the perceptual score $S$, the model learns the correspondence between input features and quality levels. During inference, setting $S$ to its maximum value allows the network to produce restorations of the highest feasible perceptual fidelity, effectively achieving controllable quality restoration under real-world degradations.

\noindent\textbf{Quality Prior Ensembles.}
Although existing NR-IQA methods are powerful, they may still harbor certain biases. To mitigate the impact of biases introduced by a single IQA model, we propose incorporating multiple IQA models as the final IQA prior. Specifically, we calculate the mean score of different IQA models:
\begin{equation}
  \centering
  \begin{aligned}
    S=\frac{1}{n}\sum_{i=1}^{n}s_{i},
    \label{eq:score ensemble}
  \end{aligned}
\end{equation}
where $s_{i}$ is the normalized score of the $i$-th IQA model. This ensemble method is used in both the codebook learning stage and the code prediction stage.

 \begin{table*}[tbp]
\setlength{\abovecaptionskip}{0.3cm}
  \centering
    \caption{Results on blind face restoration. The best two results are in {\color[HTML]{FF0000}\textbf{red}} and {\color[HTML]{00B0F0}\textbf{blue}} fonts, respectively.
  }  \label{tab:real_world_test}
  \vspace{-0.2cm}
  \resizebox{1.\textwidth}{!}{
\setlength{\tabcolsep}{2mm}
  \begin{tabular}{c|c|cccccccc}
\toprule
    Data.&Methods &\cellcolor{gray!40} TOPIQ-G$\uparrow$ & \cellcolor{gray!40} Musiq-G$\uparrow$ & \cellcolor{gray!40} Musiq-K$\uparrow$  & \cellcolor{gray!40} Musiq-A$\uparrow$ & \cellcolor{gray!40} Arniqa $\uparrow $ & \cellcolor{gray!40} MDFS $\downarrow$ & \cellcolor{gray!40} Q-Align$\uparrow$ & \cellcolor{gray!40} CLIP-IQA$\uparrow$\\
    \midrule
    \multirow{8}{*}{\rotatebox[origin=c]{90}{\textit{LFW-Test}}}
          &RestoreFormer\citep{wang2022restoreformer}&0.793&  0.807 &73.70 &4.65 &0.711&18.64&4.22&0.741\\
          &DR2\citep{wang2023dr2}&0.720 &0.753 &67.13 &4.67 &0.700&18.43&4.23&0.658\\
          &CodeFormer\citep{codeformer} &0.809&0.832& 75.47 &4.76 &0.726 &18.46&4.31&0.697\\
          &DifFace\citep{yue2022difface} &0.718&0.748&69.90&4.48&0.686&18.41&3.86&0.610\\
          &DAEFR\citep{tsai2023dual}&0.814&0.827 & 75.84&4.81&0.742&\color[HTML]{00B0F0}{\textbf{17.91}}&4.33&0.696\\
&WaveFace\citep{miao2024waveface}&0.786&0.799&73.55&4.69&0.689&19.18&4.43&\color[HTML]{00B0F0}{\textbf{0.788}}\\
          &Interlcm\citep{li2025interlcm}&\color[HTML]{00B0F0}{\textbf{0.831}}&\color[HTML]{00B0F0}\textbf{0.834}&\color[HTML]{00B0F0}\textbf{75.87}&\color[HTML]{00B0F0}\textbf{4.84}&\color[HTML]{00B0F0}{\textbf{0.753}}&17.98&\color[HTML]{00B0F0}\textbf{4.55}&0.721\\   
\rowcolor{gray!10}  \cellcolor{white}          & IQPIR (Ours) &\textcolor{red}{\textbf{0.861}}&\textcolor{red}{\textbf{0.878}}&\textcolor{red}{\textbf{76.89}}&\textcolor{red}{\textbf{5.08}}&\textcolor{red}{\textbf{0.761}}&\color[HTML]{FF0000}\textbf{17.82}&\color[HTML]{FF0000}{\textbf{4.67}}&\color[HTML]{FF0000}{\textbf{0.790}}\\
  \midrule
  \multirow{8}{*}{\rotatebox[origin=c]{90}{\textit{WebPhoto-Test}}}
          &RestoreFormer\citep{wang2022restoreformer}&0.706& 0.721 &69.83 &4.55 &0.677&19.21&3.52&0.711 \\
          &DR2\citep{wang2023dr2}&0.621 & 0.630&61.28 &\color[HTML]{00B0F0}{\textbf{4.88}}&0.601&20.51&2.87&0.447\\
&CodeFormer\citep{codeformer}&0.756 &0.782 &73.56 &4.69&0.687&19.10&3.84&0.691 \\
          &DifFace\citep{yue2022difface} &0.638&0.670&65.77&4.44&0.642&19.36&3.31&0.586\\
          &DAEFR\citep{tsai2023dual}&0.746 &0.753 &72.70 &4.58 &0.701&18.63&3.82&0.669\\
&WaveFace\citep{miao2024waveface}&0.694&0.704&70.46&4.48&0.686&19.66&3.76&\color[HTML]{00B0F0}{\textbf{0.778}}\\
&Interlcm\citep{li2025interlcm}&\color[HTML]{00B0F0}{\textbf{0.794}}&\color[HTML]{00B0F0}{\textbf{0.807}}&\color[HTML]{00B0F0}{\textbf{75.82}}&4.81&\color[HTML]{00B0F0}{\textbf{0.731}}&\color[HTML]{00B0F0}\textbf{18.45}&\color[HTML]{00B0F0}{\textbf{3.99}}&0.752\\
\rowcolor{gray!10}  \cellcolor{white}          & IQPIR (Ours) &\textcolor{red}{\textbf{0.822}}&\textcolor{red}{\textbf{0.849}}&\textcolor{red}{\textbf{76.86}}&\textcolor{red}{\textbf{4.96}}&\textcolor{red}{\textbf{0.742}}&\textcolor{red}{\textbf{18.30}}&\textcolor{red}{\textbf{4.23}}&\color[HTML]{FF0000}{\textbf{0.783}}\\
  \midrule
  \multirow{8}{*}{\rotatebox[origin=c]{90}{\textit{WIDER-Test}}} 
          &RestoreFormer\citep{wang2022restoreformer}&0.714 & 0.733& 67.83&4.49 &0.691 &19.04&3.55&0.727\\
          &DR2\citep{wang2023dr2}& 0.731&0.758 &67.76 &4.73 &0.601 &20.06&2.87&0.523\\
          &CodeFormer\citep{codeformer} &0.772&0.810&72.97 &4.57&0.705&18.67&4.05&0.697 \\
          &DifFace\citep{yue2022difface} &0.684&0.715&65.04&4.36&0.644&18.88&3.60&0.597\\
          &DAEFR\citep{tsai2023dual}&0.787 &0.807&74.15  &4.60 &0.721&\color[HTML]{00B0F0}\textbf{18.18}&4.17&0.697\\
&WaveFace\citep{miao2024waveface}&0.751&0.778&72.90&4.70&0.686&19.81&4.12&\color[HTML]{00B0F0}{\textbf{0.781}}\\
&Interlcm\citep{li2025interlcm}&\color[HTML]{00B0F0}\textbf{0.798}&\color[HTML]{00B0F0}\textbf{0.820}&\color[HTML]{00B0F0}{\textbf{75.34}}&\color[HTML]{00B0F0}{\textbf{4.80}}&\color[HTML]{00B0F0}{\textbf{0.743}}&18.26&\color[HTML]{00B0F0}\textbf{4.24}&0.754\\
\rowcolor{gray!10}   \cellcolor{white}   & IQPIR (Ours) &\textcolor{red}{\textbf{0.837}}&\textcolor{red}{\textbf{0.872}}&\textcolor{red}{\textbf{76.38}}&\textcolor{red}{\textbf{4.93}}&\textcolor{red}{\textbf{0.748}}&\color[HTML]{FF0000}{\textbf{18.06}}&\textcolor{red}{\textbf{4.56}}&\color[HTML]{FF0000}{\textbf{0.785}}\\
\bottomrule
  \end{tabular}
  }
  \vspace{-0.3cm}
\end{table*}

\subsection{Quality Optimization}
\label{3.3}
Using NR-IQA measures as objectives in image processing systems represents a promising but underexplored area. However, the optimization direction of NR-IQA objectives is unstable due to the absence of a reference. This can lead to the generation of images with low perceptual quality (or known as unrealistic) from a human perspective, known as \textbf{adversarial examples}, that nonetheless deceive NR-IQA models into assigning high scores. This phenomenon is known as over-optimization, a problem also observed in large-scale reward learning systems~\cite{gao2023scaling}.

The presence of adversarial examples renders NR-IQA objectives unreliable. A fundamental solution involves reducing the presence of these examples within the output space. We demonstrate that a discrete codebook prior naturally supports this by restricting the output space to a finite set, significantly reducing the number of adversarial examples. Training the codebook on HQ images ensures that most samples within this finite space possess high-quality semantic information, further limiting adversarial examples. Thus, NR-IQA measures are well-suited as objectives for discrete codebook-based network structures. Based on this, we propose to maximize NR-IQA scores directly to fine-tune our model's parameters.

\noindent\textbf{Final training objective.} 
In the second stage, three losses are utilized to train the encoder $\textit{E}$ and the transformer $\textit{T}$: 
\begin{equation}
  \centering
  \begin{aligned}
      \mathcal{L}_{feat}=\|Z_l-sg(Z_q)\|_2^2, 
    \label{eq:loss_lfeat2}
  \end{aligned}
\end{equation}
\begin{equation}
  \centering
  \begin{aligned}
      \mathcal{L}_{index}= \sum_{i=0}^{h*w-1}-\textbf{c}_1^i log(\widehat{\textbf{c}}_1^i) +\sum_{i=0}^{h*w-1}-\textbf{c}_2^i log(\widehat{\textbf{c}}_2^i),
    \label{eq:loss_lindex}
  \end{aligned}
\end{equation}
\begin{equation}
  \centering
  \begin{aligned}
\mathcal{L}_{quality}=-IQA(x_{res}),
    \label{eq:loss_quality}
  \end{aligned}
\end{equation}
where $\mathcal{L}_{feat} $ denotes mean square error, $\mathcal{L}_{index}$ is a cross-entropy loss, and $\mathcal{L}_{quality}$ is a quality loss. $sg(\cdot)$ denotes the stop-gradient operation. 
$\mathcal{L}_{index}$ takes the HQ code indices $\textbf{c}_1$ and $\textbf{c}_2$ from the two codebooks as supervision targets, guiding \textit{T} to predict the corresponding indices and thereby reconstruct the correct HQ representations during decoding.
For $\mathcal{L}_{feat}$, the ground-truth $Z_q$ is obtained via:
\begin{equation}
  \centering
  \begin{aligned}
      Z_q=Z_q^1+\alpha Z_q^2,
    \label{eq:stage2_gt}
  \end{aligned}
\end{equation}
where $Z_q^1$ and $Z_q^2$ are retrieved from the two codebooks according to the predicted indices $\textbf{c}_1$ and $\textbf{c}_2$.
$Z_q$ is computed following the same manner in~\cref{eq:dual_codebook} when $S>S_{thr}$ (\textit{i.e.}, for HQ+ images). Consequently, the feature reconstruction loss  $\mathcal{L}_{feat}$ guides the network to align the LQ feature representation with its corresponding HQ+ counterpart. 

Leveraging the inherent constraint of the discrete codebook prior, which restricts the output space and mitigates the generation of adversarial artifacts, we further introduce a perceptual enhancement term based on NR-IQA scores. This perception-based objective is formulated as:
\begin{equation}
    \mathcal{L}_{quality}=-IQA(x_{res}), 
\end{equation}
which encourages the model to progressively refine perceptual quality throughout optimization.

The final training objective is therefore expressed as:  
\begin{equation}
  \centering
  \begin{aligned}
      \mathcal{L}_{f}=\mathcal{L}_{feat}+\lambda_1 \mathcal{L}_{index}+\lambda_2 \mathcal{L}_{quality}, 
    \label{eq:loss_stage2}
  \end{aligned}
\end{equation}
where $\lambda_1$ and $\lambda_2$ are set to 0.5 and 0.1, respectively. This formulation enables IQPIR to balance reconstruction accuracy and perceptual refinement by directly aligning optimization with perceptual quality metrics.

\begin{table*}[t]
\centering
\setlength{\abovecaptionskip}{0cm}
\caption{Results on low-light image enhancement. 
}\label{table:low-light}
\resizebox{\textwidth}{!}{
\setlength{\tabcolsep}{1.6mm}
\begin{tabular}{l|c|cccc|cccc|cccc}
\toprule
                          &           & \multicolumn{4}{c|}{\textit{LOL-v1}}& \multicolumn{4}{c|}{\textit{LOL-v2-real}}& \multicolumn{4}{c}{\textit{LOL-v2-synthetic}}\\ \cline{3-14}
\multirow{-2}{*}{Methods} & \multirow{-2}{*}{Sources} 
& \cellcolor{gray!40}PSNR~$\uparrow$  & \cellcolor{gray!40}SSIM~$\uparrow$  
& \cellcolor{gray!40}FID~$\downarrow$ & \cellcolor{gray!40}BIQE~$\downarrow$ 
& \cellcolor{gray!40}PSNR~$\uparrow$  & \cellcolor{gray!40}SSIM~$\uparrow$  
& \cellcolor{gray!40}FID~$\downarrow$ & \cellcolor{gray!40}BIQE~$\downarrow$ 
& \cellcolor{gray!40}PSNR~$\uparrow$  & \cellcolor{gray!40}SSIM~$\uparrow$  
& \cellcolor{gray!40}FID~$\downarrow$ & \cellcolor{gray!40}BIQE~$\downarrow$ \\ \midrule
CUE~\citep{zheng2023empowering} & ICCV23 & 21.86 & 0.841 & 69.83 & 27.15 & 21.19 & 0.829 & 67.05 & 28.83 & 24.41 & 0.917 & 31.33 & 33.83 \\
GSAD~\citep{jinhui2023global} & NIPS23 & 23.23 & 0.852 & 51.64 & 19.96 & 20.19 & 0.847 & 46.77 & 28.85 & 24.22 & 0.927 & 19.24 & 25.76 \\
AST~\citep{Zhou_2024_CVPR} & CVPR24 & 21.09 & 0.858 & 87.67 & 21.23 & 21.68 & 0.856 & 91.81 & 25.17 & 22.25 & 0.927 & 37.19 & 28.78 \\
MambaIR~\citep{guo2024mambair} & ECCV24 & 22.23 & 0.863 & 63.39 & 20.17 & 21.15 & 0.857 & 56.09 & 24.46 & 25.75 & 0.937 & 19.75 & 20.37 \\
Reti-Diff~\cite{he2025reti} & ICLR25 & {\color[HTML]{00B0F0}\textbf{25.35}} & 0.866 & 49.14 & 17.75 & 22.97 & 0.858 & {\color[HTML]{00B0F0}\textbf{43.18}} & 23.66 & {\color[HTML]{00B0F0}\textbf{27.53}} &\color[HTML]{00B0F0} \textbf{0.951} & {\color[HTML]{00B0F0}\textbf{13.26}} &\color[HTML]{00B0F0} \textbf{15.77} \\
CIDNet~\cite{yan2024you} & CVPR25 & 23.50 & {\color[HTML]{00B0F0}\textbf{0.900}} & {\color[HTML]{00B0F0}\textbf{46.69}} & {\color[HTML]{00B0F0}\textbf{14.77}} & {\color[HTML]{00B0F0}\textbf{24.11}} &\color[HTML]{00B0F0} \textbf{0.871} & 48.04 &\color[HTML]{00B0F0} \textbf{18.45} & 25.71 & 0.942 & 18.60 & 15.87 \\
\rowcolor{gray!10} IQPIR & Ours & \color[HTML]{FF0000}\textbf{25.72} &\color[HTML]{FF0000} \textbf{0.902} &\color[HTML]{FF0000} \textbf{40.18} &\color[HTML]{FF0000} \textbf{13.32} &\color[HTML]{FF0000} \textbf{24.20} &\color[HTML]{FF0000} \textbf{0.889} &\color[HTML]{FF0000} \textbf{40.05} &\color[HTML]{FF0000} \textbf{13.39} &\color[HTML]{FF0000} \textbf{27.60} &\color[HTML]{FF0000} \textbf{0.953} &\color[HTML]{FF0000} \textbf{12.04} &\color[HTML]{FF0000} \textbf{15.30}
\\ \bottomrule
\end{tabular}}
\vspace{-3mm}
\end{table*}

\section{Experiments}
\noindent\textbf{Implementation Details}.
\label{sec:4.1}
IQPIR is implemented in PyTorch and trained on one NVIDIA H200 GPU.
The quantized feature maps have dimensions of $16 \times 16 \times 256$, and each codebook consists of 1024 entries, each with a 256-dimensional embedding. We adopt Adam with momentum terms of (0.9,0.999) and set the batch size to 32.

For our quality priors, we employ an ensemble of three advanced NR-IQA models, incorporating traditional and MLLM-based methods to estimate the final quality score.
{For the face task, this ensemble specifically integrates two traditional approaches (Musiq-GFIQA and TOPIQ-GFIQA, both trained on GFIQA\cite{su2023going}) with one MLLM-based method (Q-Align \cite{wu2023q}). 
In other tasks, quality assessment is performed using  Musiq\cite{ke2021musiq}, CLIP-IQA\cite{wang2023exploring}, and BRISQUE\cite{mittal2012no}.}

\begin{table}[t]
\begin{minipage}[c]{0.5\textwidth}
\centering
\setlength{\abovecaptionskip}{0cm}
\setlength{\belowcaptionskip}{0.05cm}
\caption{Results on underwater image enhancement.}
\resizebox{\columnwidth}{!}{
\setlength{\tabcolsep}{1.7mm}
\begin{tabular}{l|c|cccc}
\toprule
 && \multicolumn{4}{c}{\textit{UIEB}}\\ \cline{3-6}
\multirow{-2}{*}{Methods} &\multirow{-2}{*}{Sources} & \cellcolor{gray!40}PSNR~$\uparrow$ & \cellcolor{gray!40}SSIM~$\uparrow$& \cellcolor{gray!40}UCIQE~$\uparrow$& \cellcolor{gray!40}UIQM~$\uparrow$\\ \midrule
PUGAN~\citep{cong2023pugan}& TIP23& { {23.05}} & 0.897                                 & 0.608                                 & 2.902                                          \\
ADP~\citep{zhou2023underwater}& IJCV23   & 22.90                                 & 0.892                                 &0.621 & 3.005                                  \\
NU2Net~\citep{guo2023underwater}   & AAAI23 & 22.38                                 & 0.903                                 & 0.587                                 & 2.936                                 \\
AST~\citep{Zhou_2024_CVPR}  & CVPR24 & 22.19 & 0.908 & 0.602 & 2.981 \\
MambaIR~\citep{guo2024mambair} &ECCV24 &22.60 &{\color[HTML]{00B0F0} \textbf{0.916}} &{{0.617}} &2.991 \\
Reti-Diff~\cite{he2025reti}              & ICLR25        & {\color[HTML]{00B0F0} \textbf{24.12}} & {{0.910}} & {\color[HTML]{00B0F0} \textbf{0.631}} & {\color[HTML]{00B0F0} \textbf{3.088}} \\ 
\rowcolor{gray!10} IQPIR & Ours & \color[HTML]{FF0000} \textbf{24.35} &\color[HTML]{FF0000} \textbf{0.918} &\color[HTML]{FF0000} \textbf{0.642} &\color[HTML]{FF0000} \textbf{3.127} \\ \bottomrule
\end{tabular}}
\label{table:Underwater} \vspace{-6mm}
\end{minipage}
\end{table}

\begin{figure*}[htbp]
  \setlength{\abovecaptionskip}{-0.3cm}
  \centering
    \includegraphics[scale=0.3]{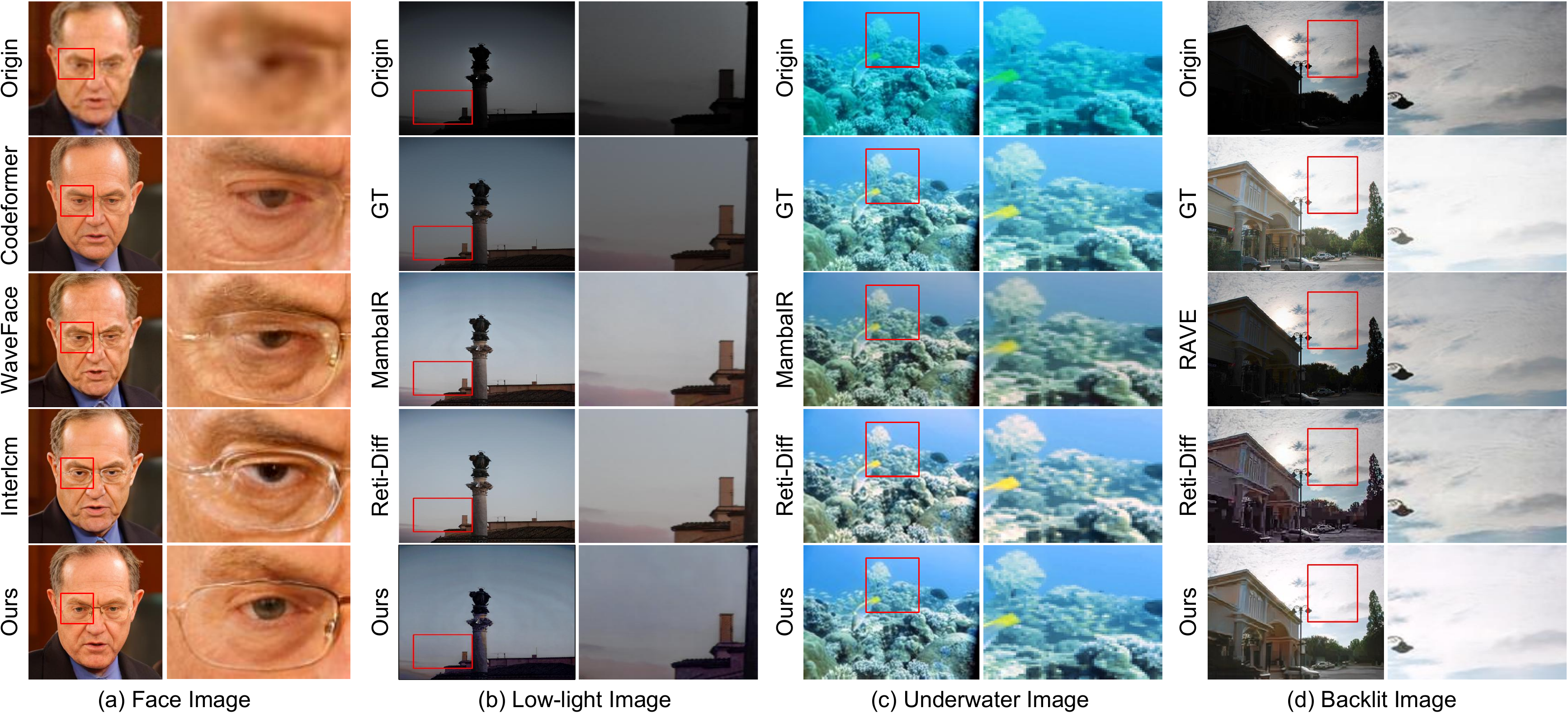}
    \newline
    \caption{Visualizations on low light, underwater and backlit image restoration.}
    \label{Fig.LOL_backlit_underwater}
\vspace{-0.7cm}
\end{figure*}

\subsection{Comparative Evaluation}
\label{sec:4.3}
\noindent\textbf{Blind face restoration}. Following the common setting~\cite{codeformer,yue2022difface}, we train our model on \textit{FFHQ} and evaluate the performance on three real-world data: LFW-Test \citep{huang2008labeled}, WebPhoto-Test \citep{wang2021gfpgan}, and WIDER-Test \citep{codeformer}. We apply 8 NR-IQA metrics for evaluation, including Musiq-Koniq \citep{ke2021musiq}, Musiq-GFIQA \citep{ke2021musiq}, and TOPIQ-GFIQA \citep{chen2023topiq},  Musiq-AVA \citep{ke2021musiq} (for aesthetic quality), Arniqa \citep{agnolucci2024arniqa}, MDFS \citep{ni2024opinion}, CLIP-IQA \citep{wang2023exploring} and Q-Align \citep{wu2023q}. 
Results in~\cref{tab:real_world_test} show that IQPIR achieves superior performance across all datasets.
Particularly, for Musiq-GFIQA and TOPIQ-GFIQA (metrics designed for evaluating face quality), IQPIR attains substantial improvements.
Qualitative comparisons in \cref{Fig.LOL_backlit_underwater} further demonstrate that IQPIR recovers finer facial details with more realistic textures and natural color tones.

\noindent\textbf{Low-light image enhancement}. Following the setting of Reti-Diff~\cite{he2025reti}, we evaluate the performance on \textit{LOL-v1}~\cite{wei2018deep}, \textit{LOL-v2-real}~\cite{yang2021sparse}, and \textit{LOL-v2-synthetic}~\cite{yang2021sparse}.
Four metrics are used for evaluation, including PSNR, SSIM, FID \cite{heusel2017gans}, and BIQE \cite{moorthy2010two}, where higher PSNR or SSIM, and lower FID or BIQE correspond to better results. As depicted in~\cref{table:low-light}, our IQPIR achieves a leading place across all metrics, indicating the superiority of our perception prior.

\begin{table}[t]
\begin{minipage}[c]{0.5\textwidth}
\centering
\setlength{\abovecaptionskip}{0cm}
\setlength{\belowcaptionskip}{0.05cm}
\caption{Results on backlit image enhancement.}
\resizebox{\columnwidth}{!}{
\setlength{\tabcolsep}{2mm}
\begin{tabular}{l|c|cccc}
\toprule
&  & \multicolumn{4}{c}{\textit{BAID}}                                                                                                                             \\ \cline{3-6}
\multirow{-2}{*}{Methods} & \multirow{-2}{*}{Sources} & \cellcolor{gray!40}PSNR~$\uparrow$ & \cellcolor{gray!40}SSIM~$\uparrow$ & \cellcolor{gray!40}LPIPS~$\downarrow$ & \cellcolor{gray!40}FID~$\downarrow$ \\ \midrule
CLIP-LIT~\citep{liang2023iterative}                  & ICCV23                      & 21.13                                 & 0.853                                 &0.159 &37.30 \\
DiffIR~\cite{xia2023diffir}                    & ICCV23                      & 21.10                                 & 0.835                                 & 0.175                                 & 40.35                                 \\
AST~\cite{Zhou_2024_CVPR}  & CVPR24 &22.61 & 0.851 & 0.156 & 32.47 \\
MambaIR~\cite{guo2024mambair} &ECCV24 &{{23.07}} &{{0.874}} &{{0.153}} &{{29.13}} \\
RAVE~\cite{gaintseva2024rave} & ECCV24 & 21.26 & 0.872 & \color[HTML]{00B0F0} \textbf{0.096} & 64.89 \\
Reti-Diff~\cite{he2025reti}  & ICLR25                    & {\color[HTML]{00B0F0} \textbf{23.19}} & {\color[HTML]{00B0F0} \textbf{0.876}} & {{0.147}} & {\color[HTML]{00B0F0} \textbf{27.47}} \\
\rowcolor{gray!10} IQPIR & Ours &\color[HTML]{FF0000} \textbf{24.06} &\color[HTML]{FF0000} \textbf{0.885} &\color[HTML]{FF0000} \textbf{0.083} &\color[HTML]{FF0000} \textbf{25.32} \\ \bottomrule
\end{tabular}}
\label{table:backlit} \vspace{-6mm}
\end{minipage}
\end{table}

\begin{table*}[th!]
\begin{minipage}[c]{\textwidth}
  \caption{Break down ablation.}
  \vspace{-0.3cm}
\centering
  \resizebox{\linewidth}{!}{
  \setlength{\tabcolsep}{2.3mm}
  \begin{tabular}{c|c|c|c|c|c|c|c|c|c|c}
    \toprule
    ID&  \makecell[c]{Score\\ Condition}&\makecell[c]{Dual \\ Codebook}&\makecell{ Quality \\ Optimization}& TOPIQ-G $\uparrow$  & Musiq-K $\uparrow$ & Arniqa $\uparrow$ & MDFS $\downarrow$ & Q-Align $\uparrow$ & CLIP-IQA $\uparrow$  &  \makecell[c]{Inference\\ time (\textbf{ms})} $\downarrow$\\
    \midrule
    (a)&$\times$ &$\times$ &$\times$ &0.757&73.53 &0.675 & 20.03 &3.78 & 0.742 & 5.27\\
   (b) & $\checkmark$ &$\times$ &$\times$  &0.790&75.95 &0.713 & 18.87 &4.03 & 0.768 & 5.39\\
   (c) & $\checkmark$ & $\checkmark$ & $\times$ &0.806 &76.04 & 0.731 & 18.52  &4.16 & 0.776 & 5.63\\
   (d)  & $\checkmark$ & $\checkmark$ &$\checkmark$ &\textbf{0.822}&\textbf{76.86} & \textbf{0.742} & 18.30 & \textbf{4.23} & 0.783  & 5.63 \\
    \bottomrule
  \end{tabular}
  }
  \label{tab:ablation}
\end{minipage}\\
\begin{minipage}[c]{\textwidth}
\centering
\setlength{\abovecaptionskip}{0.05cm}
\caption{User study, where we conduct experiments on blind face restoration and low-light image enhancement.
}
		\label{table:userstudy} 
\resizebox{\textwidth}{!}{
\setlength{\tabcolsep}{2.1mm}
    \begin{tabular}{l cccc c l cccc}
        \toprule
        \multicolumn{5}{c}{(a) Blind Face Restoration } & & \multicolumn{5}{c}{(b) Low-light Image Enhancement} \\
        \cmidrule(lr){1-5} \cmidrule(lr){7-11}
        Methods & \makecell[c]{\textit{LFW Test}}& \makecell[c]{\textit{WebPhoto Test}} & \makecell[c]{\textit{WIDER Test}} & Mean & & Methods & \textit{LOL-v1} & \makecell[c]{\textit{LOL-v2-real}} & \makecell[c]{\textit{LOL-v2-synthetic}} & Mean\\
          \cmidrule(lr){1-5} \cmidrule(lr){7-11}
        DifFace~\cite{yue2022difface} & 2.50 & 2.58 & 2.08 & 2.39 & & AST~\cite{Zhou_2024_CVPR} & 3.08 & 3.42 & 3.58 & 3.36 \\
        DAEFR~\cite{tsai2023dual} & 3.42 & 3.58 & 2.17 & 3.06 & & MambaIR~\cite{guo2024mambair} & 3.67 & 3.58 & 3.16 & 3.47\\
        WaveFace~\cite{miao2024waveface} & 3.92 & 4.08 & 3.17 & 3.72 & & Reti-Diff~\cite{he2025reti} & 4.42 & 3.50 & 3.67 & 3.86\\
        Interlcm~\cite{li2025interlcm} & 3.58 & 3.92 & 3.42 & 3.64 & & CIDNet~\cite{yan2024you} & 3.50 & 3.25 & 3.59 & 3.45\\
        \rowcolor{gray!10} IQPIR (Ours) & \textbf{4.17} & \textbf{4.25} & \textbf{3.67} & \textbf{4.03} & & IQPIR (Ours) & \textbf{4.58} & \textbf{3.92} & \textbf{4.17} & \textbf{4.22}\\
        \bottomrule
    \end{tabular}}
\vspace{-3mm}
\end{minipage}
\end{table*}

\begin{figure*}[t]
\begin{minipage}[c]{0.52\textwidth}
\centering
\setlength{\abovecaptionskip}{0cm}
\setlength{\belowcaptionskip}{0.05cm}
\includegraphics[width=1\linewidth]{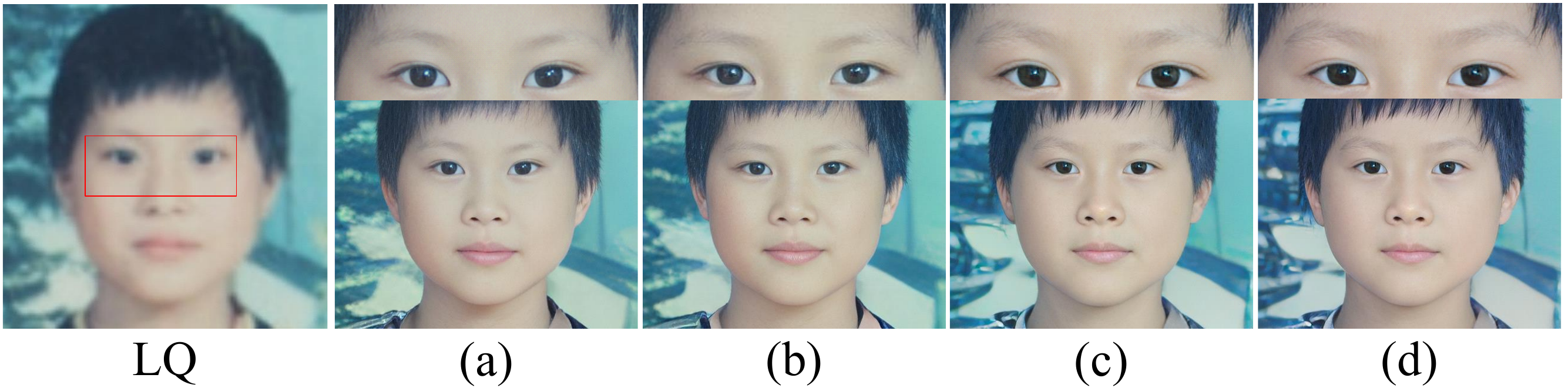}   
	\caption{Visualization of breakdown ablation, where (a), (b), (c), and (d) are consistent with those in \cref{tab:ablation}.}
    \label{Fig.ablation_vis}
    \vspace{-0.5cm}
\end{minipage}
\begin{minipage}[c]{0.47\textwidth}
\centering
\setlength{\abovecaptionskip}{0cm}
\setlength{\belowcaptionskip}{0.05cm}
\includegraphics[width=1\linewidth]{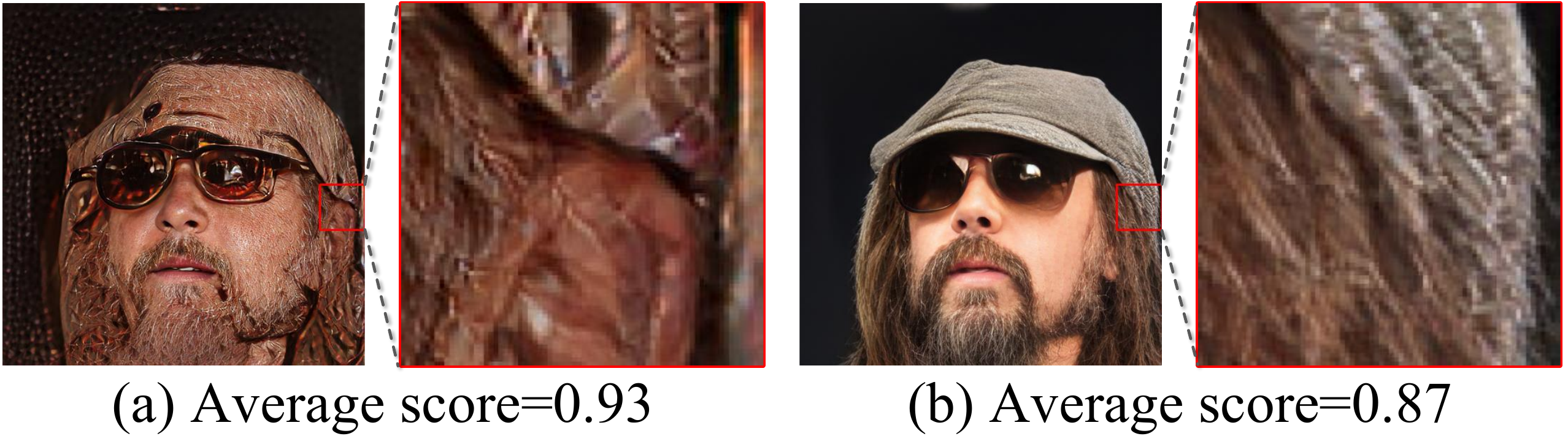}   
	\caption{Quality in continuous (a) and discrete (b) representation space. Our discrete strategy shows score-quality consistency.}
    \label{Fig.discrete_continuous}
    \vspace{-0.5cm}
\end{minipage} \\
\end{figure*}

\noindent\textbf{Underwater image enhancement}. Following~\cite{he2025reti}, we evaluate the performance on the \textit{UIEB}~\cite{li2019underwater} dataset. Two extra metrics, UCIQE~\cite{yang2015underwater} and UIQM~\cite{panetta2015human}, are used, where higher values mean better performance. As shown in~\cref{table:Underwater,Fig.LOL_backlit_underwater}, we achieve better performance qualitatively and quantitatively. This demonstrate that IQPIR effectively improves image quality under underwater conditions, outperforming competing methods in clarity and visual appeal.

\noindent\textbf{Backlit image enhancement}. Following \cite{liang2023iterative}, we evaluate the performance of our method on the \textit{BAID} dataset using PSNR, SSIM, LPIPS~\cite{zhang2018unreasonable}, and FID metrics. A lower LPIPS value indicates better perceptual quality. As shown in~\cref{table:backlit,Fig.LOL_backlit_underwater}, our approach achieves leading results, demonstrating clear advantages in detail preservation and reduced color distortion compared to competing methods.

\subsection{Ablation Study}
\vspace{-0.15cm}
{We conduct experiments on \textit{WebPhoto-Test}.}
For space limitations, only partial ablations are included; see supplementary materials for more details, such as the size of HQ+ Codebook, the fusion manners of quantized feature, and the incorporation manner of the perception prior.

\noindent\textbf{Quality Prior Conditional Approach}.
We typically use the maximum score as the condition to achieve the best restoration quality. As shown in \cref{tab:ablation}, (a) and (b) demonstrate the effectiveness of incorporating the quality prior into the codebook lookup Transformer. We further compare per-image runtime, showing that our approach achieves performance gains with only marginal computational overhead.
The visualization in \cref{Fig.ablation_vis} also supports this—results from (b) produce sharper edges and richer textures compared to (a), validating the benefit of perception priors.

\noindent\textbf{Effectiveness of Dual Codebook}.
The comparison between experiments (b) and (c) in \cref{tab:ablation} reveals that the proposed dual codebook consistently improves perceptual fidelity over the traditional single-codebook baseline.
\cref{Fig.ablation_vis} further illustrates that (b) generates more distinct facial details than (c), confirming the capability of the HQ+ codebook to preserve high-quality features.

\noindent\textbf{Quality Optimization}.
Ablations (c) and (d) in \cref{tab:ablation} highlight the effectiveness of quality optimization. The discrete representation learned by the codebook enables the model to enhance reconstruction quality while avoiding the over-optimization issue often seen in continuous spaces.
As depicted in \cref{Fig.ablation_vis}, quality optimization further refines details and improves overall perceptual realism.

\begin{table*}[t]
\begin{minipage}[c]{0.72\columnwidth}
\centering
\setlength{\abovecaptionskip}{0cm}
\setlength{\belowcaptionskip}{0.05cm}
\caption{Effect of our IQPIR (BFR).}
\resizebox{\columnwidth}{!}{
\setlength{\tabcolsep}{0.6mm}
\begin{tabular}{l|c|cccc}
\toprule
Data.                         & Metrics & WavwFace  &\cellcolor{gray!10} WaveFace+  & InterIcm  &\cellcolor{gray!10} InterIcm+ \\ \midrule
\multirow{5}{*}{\rotatebox[origin=c]{90}{\textit{LFW-Test}}} & Topiq-G~$\uparrow$    & 0.786    & \cellcolor{gray!10}0.802  & 0.831 & \cellcolor{gray!10} 0.850           \\
& Musiq-G~$\uparrow$   & 0.799   & \cellcolor{gray!10}0.817     & 0.834     & \cellcolor{gray!10} 0.852  \\
& Q-Align~$\uparrow$    &4.43    &\cellcolor{gray!10} 4.53    & 4.55 & \cellcolor{gray!10} 4.67 \\
&CLIP-IQA~$\uparrow$   & 0.788 & \cellcolor{gray!10} 0.793 & 0.721 & \cellcolor{gray!10} 0.737 \\
                                 &Gain    & ---   & \cellcolor{gray!10} 1.8 $\%$  &--- & \cellcolor{gray!10} 2.33$\%$ \\ \midrule
\multirow{5}{*}{\rotatebox[origin=c]{90}{\textit{WIDER-Test}}}& Topiq-G~$\uparrow$    & 0.751    & \cellcolor{gray!10}0.775  & 0.798 & \cellcolor{gray!10} 0.813           \\
                                 & Musiq-G~$\uparrow$   & 0.778   & \cellcolor{gray!10}0.808     & 0.820     & \cellcolor{gray!10} 0.844  \\
                                 & Q-Align~$\uparrow$    &4.12    &\cellcolor{gray!10} 4.35    & 4.24 & \cellcolor{gray!10} 4.38 \\
                                 &CLIP-IQA~$\uparrow$   & 0.781 & \cellcolor{gray!10} 0.792 & 0.754 & \cellcolor{gray!10} 0.765 \\
                                 &Gain    & ---   & \cellcolor{gray!10} 3.54 $\%$  &--- & \cellcolor{gray!10} 2.39$\%$  \\ \bottomrule         
\end{tabular}}
\label{table:IQPIR_BFR}
\end{minipage}
\begin{minipage}[c]{0.625\columnwidth}
\centering
\setlength{\abovecaptionskip}{0cm}
\setlength{\belowcaptionskip}{0.05cm}
\caption{Effect of our IQPIR (LLIE).}
\resizebox{\columnwidth}{!}{
\setlength{\tabcolsep}{0.6mm}
\begin{tabular}{l|c|cccc}
\toprule
Data.                         & Metrics & Reti-Diff  &\cellcolor{gray!10} Reti-Diff+  & CIDNet  &\cellcolor{gray!10} CIDNet+ \\ \midrule
\multirow{5}{*}{\rotatebox[origin=c]{90}{\textit{LOL-v1}}} & PSNR~$\uparrow$    & 25.35    & \cellcolor{gray!10}25.52  & 23.50 & \cellcolor{gray!10} 24.28           \\
                                 & SSIM~$\uparrow$   & 0.866   & \cellcolor{gray!10}0.889     & 0.900     & \cellcolor{gray!10} 0.908  \\
                                 & FID~$\downarrow$    &49.14    &\cellcolor{gray!10} 42.15    & 46.69 & \cellcolor{gray!10} 40.36 \\
                                 &BIQE~$\downarrow$   & 17.75 & \cellcolor{gray!10} 15.35 & 14.77 & \cellcolor{gray!10} 13.73 \\
                                 &Gain    & ---   & \cellcolor{gray!10} 7.77 $\%$  &--- & \cellcolor{gray!10} 6.20$\%$ \\ \midrule
\multirow{5}{*}{\rotatebox[origin=c]{90}{\textit{LOL-v2-real}}}& PSNR~$\uparrow$    & 22.97    & \cellcolor{gray!10}23.73  & 24.11 & \cellcolor{gray!10} 24.73           \\
                                 & SSIM~$\uparrow$   & 0.858   & \cellcolor{gray!10}0.868     & 0.871     & \cellcolor{gray!10} 0.882  \\
                                 & FID~$\downarrow$   &43.18    &\cellcolor{gray!10} 40.15    & 48.04 & \cellcolor{gray!10} 43.33 \\
                                 &BIQE~$\downarrow$   & 23.66 & \cellcolor{gray!10} 19.57 & 18.45 & \cellcolor{gray!10} 16.31 \\
                                 &Gain    & ---   & \cellcolor{gray!10} 7.19 $\%$  &--- & \cellcolor{gray!10} 5.94$\%$  \\ \bottomrule         
\end{tabular}}
\label{table:IQPIR_LLIE}
\end{minipage}
\begin{minipage}[c]{0.685\columnwidth}
\centering
\setlength{\abovecaptionskip}{0cm}
\setlength{\belowcaptionskip}{0.06cm}
\caption{Effect of our IQPIR (UIE and BIE).}
\resizebox{\columnwidth}{!}{
\setlength{\tabcolsep}{0.6mm}
\begin{tabular}{l|c|cccc}\toprule
Data.                         & Metrics & MambaIR  &\cellcolor{gray!10} MambaIR+  & Reti-Diff  &\cellcolor{gray!10} Reti-Diff+ \\ \midrule
\multirow{5}{*}{\rotatebox[origin=c]{90}{\textit{UIEB}}} & PSNR~$\uparrow$    & 22.60    & \cellcolor{gray!10}23.72  & 24.12 & \cellcolor{gray!10} 24.28           \\
                                 & SSIM~$\uparrow$   & 0.914   & \cellcolor{gray!10}0.920     & 0.911     & \cellcolor{gray!10} 0.917  \\
                                 & UCIQE~$\uparrow$    &0.617    &\cellcolor{gray!10} 0.632    & 0.630 & \cellcolor{gray!10} 0.638 \\
                                 &UIQM~$\uparrow$  & 2.991 & \cellcolor{gray!10} 3.208 & 3.087 & \cellcolor{gray!10} 3.219 \\
                                 &Gain    & ---   & \cellcolor{gray!10} 3.77 $\%$  &--- & \cellcolor{gray!10} 1.69$\%$ \\ \midrule
\multirow{5}{*}{\rotatebox[origin=c]{90}{\textit{BAID}}}& PSNR~$\uparrow$    & 23.07    & \cellcolor{gray!10}23.62  & 23.19 & \cellcolor{gray!10} 23.83           \\
                                 & SSIM~$\uparrow$   & 0.874   & \cellcolor{gray!10}0.877     & 0.876     & \cellcolor{gray!10} 0.880  \\
                                 & LPIPS~$\downarrow$   &0.153    &\cellcolor{gray!10} 0.137    & 0.146 & \cellcolor{gray!10} 0.125 \\
                                 &FID~$\downarrow$   & 29.13 & \cellcolor{gray!10} 27.85 & 27.47 & \cellcolor{gray!10} 24.76 \\
                                 &Gain    & ---   & \cellcolor{gray!10} 4.37 $\%$  &--- & \cellcolor{gray!10} 7.04$\%$  \\ \bottomrule       
\end{tabular}}
\label{table:IQPIR_UIE} 
\end{minipage}\vspace{-4mm}
\end{table*}

\begin{figure*}[t]
\begin{minipage}[c]{0.395\textwidth}
\centering
\setlength{\abovecaptionskip}{0cm}
\setlength{\belowcaptionskip}{0.1cm}
\includegraphics[width=1\linewidth]{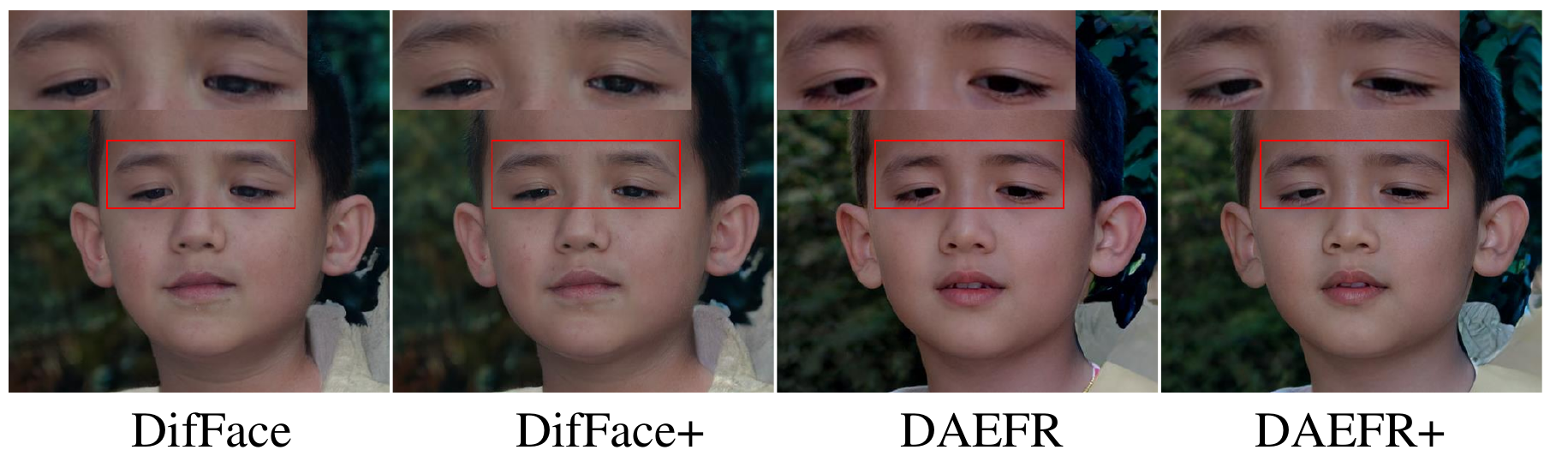}   
	\caption{Generalization of our IQPIR.}
    \label{Fig.difface+}\vspace{1mm}
\end{minipage}
\begin{minipage}[c]{0.60\textwidth}
\centering
\setlength{\abovecaptionskip}{0cm}
\setlength{\belowcaptionskip}{0.05cm}
\includegraphics[width=1\linewidth]{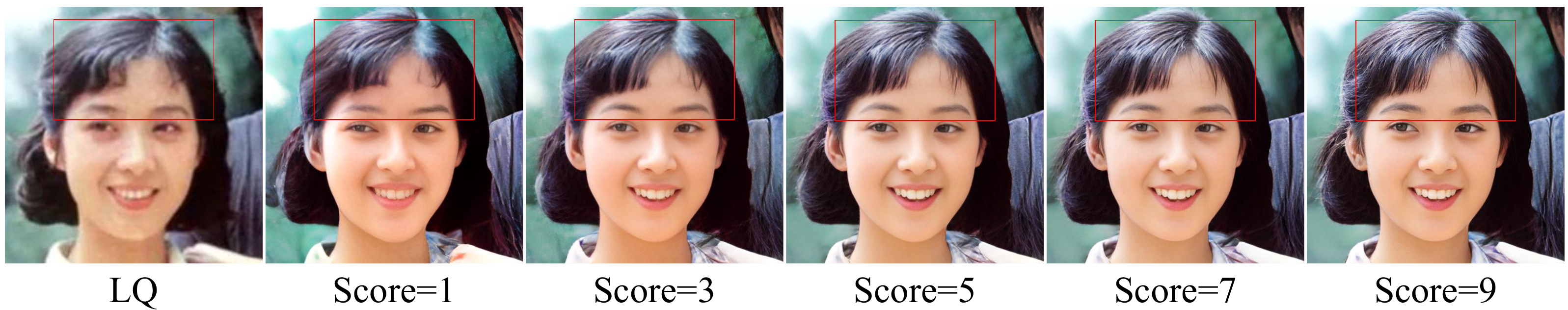} 
	\caption{Visualization with different condition scores.}
    \label{Fig.score_control} \vspace{2mm}
\end{minipage} \\ \vspace{3mm}
\begin{minipage}[t]{\textwidth}
\vspace{-2mm} 
\centering
\setlength{\abovecaptionskip}{0cm}
\captionof{table}{Low-light image detection on \textit{ExDark}.
}
		\label{table:Detection} 
\resizebox{0.795\columnwidth}{!}{
\setlength{\tabcolsep}{1.9mm}
\begin{tabular}{l|cccccccccccc|c}
\toprule
Methods (AP)& \cellcolor{c2!50}Bicycle  & \cellcolor{c2!50}Boat & \cellcolor{c2!50}Bottle  & \cellcolor{c2!50}Bus & \cellcolor{c2!50}Car & \cellcolor{c2!50}Cat  & \cellcolor{c2!50}Chair &\cellcolor{c2!50} Cup & \cellcolor{c2!50}Dog & \cellcolor{c2!50} Motor & \cellcolor{c2!50}People & \cellcolor{c2!50}Table  & \cellcolor{c2!50}Mean \\ \midrule
Baseline & 74.7                                 & 64.9                                 & 70.7                                 & 84.2                                 & 79.7                                 & 47.3                                 & 58.6                                 & 67.1                                 & 64.1                                 & 66.2                                 & 73.9                                 & 45.7                                 & 66.4                                 \\
MIRNet~\cite{zamir2020learning} & 74.9 & 69.7 & 68.3 & 89.7 & 77.6 & 57.8 & 56.9 & 66.4 & 69.7 & 64.6 & 74.6 & 53.4 & 68.6 \\ 
RUAS~\cite{liu2021retinex} & 75.7 & 71.2 & 73.5 & 90.7 & 80.1 & 59.3 & 67.0 & 66.3 & 68.3 & 66.9 & 72.6 & 50.6 & 70.2 \\
Restormer~\cite{zamir2022restormer} & 77.0 & 71.0 & 68.8 & 91.6 & 77.1 & 62.5 & 57.3 & 68.0 & 69.6 & 69.2 & 74.6 & 49.7 & 69.7 \\
SCI~\cite{ma2022toward}   & 73.4                                 & 68.0                                 & 69.5                                 & 86.2                                 & 74.5                                 & 63.1                                 & 59.5                                 & 61.0                                 & 67.3                                 & 63.9                                 & 73.2                                 & 47.3                                 & 67.2                                 \\
SNR-Net~\cite{xu2022snr}   & {{78.3}} & 74.2                                 & {{74.5}} & 89.6                                 & {{82.7}} & {{66.8}} & 66.3                                 & 62.5                                 & 74.7                                 & 63.1                                 & 73.3                                 & {{57.2}} & 71.9                                 \\
Reti-Diff~\cite{he2025reti} & 82.0 & 77.9 & {{76.4}} & 92.2 & 83.3& 69.6 & 67.4 & 74.4 & 75.5 &74.3 & \color[HTML]{00B0F0} \textbf{78.3}& {{57.9}} & 75.8\\
CIDNet~\cite{yan2024you} & 81.8 & 77.6 & 77.2 & 85.8 & 77.3 & 68.1 & 65.5 & 73.6 & 74.7 & 70.2 & 71.0 & 60.3 & 73.6 \\
\rowcolor{gray!10} Reti-Diff+ & 82.5 & \color[HTML]{00B0F0} \textbf{78.3}   & 77.2   & {\color[HTML]{FF0000} \textbf{92.7}}& {\color[HTML]{FF0000} \textbf{83.7}} & \color[HTML]{00B0F0} \textbf{70.5}   & \color[HTML]{00B0F0} \textbf{69.6} & \color[HTML]{00B0F0} \textbf{76.3} & 77.2 & \color[HTML]{00B0F0} \textbf{76.1} & {\color[HTML]{FF0000} \textbf{79.7}} & 60.2 & \color[HTML]{00B0F0} \textbf{77.0}  \\
\rowcolor{gray!10} CIDNet+    & \color[HTML]{00B0F0} \textbf{82.7} & 78.2 & {\color[HTML]{FF0000} \textbf{77.6}} & 87.6 & 77.9 & 69.700 & 67.3 & 75.9 & \color[HTML]{00B0F0} \textbf{77.3} & 73.8 & 72.5 & \color[HTML]{00B0F0} \textbf{63.7} & 75.4 \\
\rowcolor{gray!10} IQPIR & {\color[HTML]{FF0000} \textbf{83.4}} & {\color[HTML]{FF0000} \textbf{80.2}}   & \color[HTML]{00B0F0} \textbf{77.5}   & \color[HTML]{00B0F0} \textbf{92.5} & \color[HTML]{00B0F0} \textbf{83.6} & \color[HTML]{FF0000} \textbf{75.7}   & {\color[HTML]{FF0000} \textbf{77.8}} & {\color[HTML]{FF0000} \textbf{78.4}} & {\color[HTML]{FF0000} \textbf{79.3}} & {\color[HTML]{FF0000} \textbf{82.5}} & 77.3 &{\color[HTML]{FF0000} \textbf{65.3}} & {\color[HTML]{FF0000} \textbf{79.5}}
\\ \bottomrule
\end{tabular}}
\end{minipage}\vspace{-9mm} 
\end{figure*}

\noindent\textbf{Discrete Representation Mitigate Over-optimization}. As shown in \cref{Fig.discrete_continuous}, we compare quality optimization in continuous and discrete representation spaces. In the continuous space, results achieve high quality scores but exhibit poor perceptual realism. In contrast, the discrete representation space achieves both high scores and strong human-perceived quality, effectively mitigating over-optimization.

\subsection{Further Analysis and Extended Applications}
\noindent \textbf{User Study}. We conduct a user study on BFR (\textit{LFW Test}, \textit{WebPhoto Test}, and \textit{WIDER Test}) and LLIE (\textit{LOL-v1}, \textit{LOL-v2-real}, and \textit{LOL-v2-synthesize}) tasks. Twelve participants evaluated restored images on a 0–5 scale (0=worst,5=best) based on three criteria: color fidelity, noise/artifact suppression, and structural preservation. For fairness, each low-quality image and its restored one were displayed side by side, with method identities hidden. As shown in \cref{table:userstudy}, our method consistently received the highest ratings, confirming its superiority in visual quality and perception.

\noindent \textbf{The Effect of Condition Score during Testing}. We show that the condition score directly influences restoration quality during testing. As illustrated in \cref{Fig.score_control}, adjusting the target score allows continuous control over perceptual quality. In practice, we typically use the maximum score to achieve the best restoration results. Although quality controllability may seem less essential for BFR, its main purpose is to ensure the model reaches the highest-quality output rather than to serve as an independent control mechanism.

\noindent \textbf{Generalizability of IQPIR}. We further evaluate the generalization capability of IQPIR by integrating the proposed perception prior into several SOTA restoration methods across diverse tasks. For comprehensive evaluation, four specific metrics are employed for each task. As shown in \cref{table:IQPIR_BFR,table:IQPIR_LLIE,table:IQPIR_UIE}, IQPIR consistently enhances the performance of six representative models on their benchmarks, confirming the versatility and robustness of our framework. To further verify this, we add a visualization in \cref{Fig.difface+}, where the method integrated with our framework shows more detailed texture information and more harmonious visual fidelity.

\noindent\textbf{Color Enhancement}. We compare IQPIR with CodeFormer~\cite{codeformer} and CodeFormer+, where CodeFormer+ denotes CodeFormer enhanced with our perception prior.
IQPIR produces results with more natural color rendition and richer facial details than competing methods.
Moreover, CodeFormer+ yields more visually balanced and realistic color tones compared to the original CodeFormer. This indicates that our IQPIR is a plug-and-play image prior with strong generalization capacity.

\noindent\textbf{Benefits for downstream tasks}. Improving image quality can facilitate various downstream vision tasks. To verify this, we assess the impact of enhancement on low‑light object detection. Specifically, low‑quality images from ExDark~\cite{loh2019getting} are enhanced using different algorithms and subsequently fed into YOLO for detection. For reference, we also include a baseline that directly uses the original low‑quality inputs without enhancement.
As reported in \cref{table:Detection}, IQPIR achieves the highest detection performance and further provides consistent performance gains when integrated into other restoration frameworks.

\section{Limitations and Future Works}
The quality priors employed by our IQPIR are directly derived from existing NR‑IQA models.
Although this design simplifies implementation and enables effective perceptual guidance, it may also inherit biases intrinsic to the IQA models, which can lead to potential inaccuracies in quality estimation.
In future work, we plan to address this limitation by exploring more principled approaches to integrate quality priors with restoration models.
In particular, we will investigate strategies to mitigate the influence of biases introduced by individual IQA models, such as adopting adaptive ensemble schemes or jointly optimizing the IQA priors together with the restoration objectives.

\section{Conclusions}
We presented IQPIR, a perception‑driven framework for RWIR that integrates IQP with discrete codebook representations.
By leveraging NR‑IQA–based priors and a dual‑codebook architecture, IQPIR guides the restoration process toward perceptually optimal results while preserving structural fidelity.
Our quality‑conditioned transformer further enhances adaptation to diverse degradation types.
Experiments demonstrate that IQPIR achieves SOTA performance and offers a generalizable paradigm that can be readily integrated into existing restoration frameworks.

{
    \small
    \normalem
    \bibliographystyle{ieeenat_fullname}
    \bibliography{main}
}

\end{document}